\documentclass[conference]{IEEEtran}
\IEEEoverridecommandlockouts    

\usepackage{multirow}
\usepackage{lipsum}
\usepackage{amsmath}
\usepackage{amssymb}
\usepackage[ruled,vlined]{algorithm2e}
\usepackage{graphicx}
\graphicspath{{./Figures/}}
\usepackage[colorlinks=true, linkcolor=blue, citecolor=blue, urlcolor=blue]{hyperref}
\usepackage{booktabs}
\usepackage{cleveref}
\usepackage{xcolor}
\let\labelindent\relax
\usepackage{enumitem}

\title{\LARGE \bf Learning to Interact in Multi-Agent Self-Play for Controllable Autonomous Driving Scenario Generation}

\author{
	\parbox{\textwidth}{%
		\centering
		Beyazit Yalcinkaya$^{\dagger}$, Ethan Pronovost, Maximilian Naumann, Marcell Vazquez-Chanlatte%
	}%
	\thanks{Zoox, Foster City, USA
		{\tt\small \{byalcinkaya, epronovost, mnaumann, mvazquez-chanlatte\}@zoox.com}}%
	\thanks{$^{\dagger}$Work done during an internship at Zoox}%
}

\begin{document}
	
	\maketitle
	\thispagestyle{empty}
	\pagestyle{empty}
	
\begin{abstract}
	Large-scale, multi-agent self-play trains robust driving policies effectively in simulation, but controlling these models to generate specific, safety-critical edge cases remains a key challenge. Existing goal-conditioned self-play frameworks rely primarily on static map-location goals and fail to reliably produce complex agent interactions such as aggressive tailgating or adversarial cut-offs. These agent interactions are necessary for rigorous autonomous vehicle (AV) testing. To address this, we present an extended multi-agent self-play framework that introduces agent-relative dynamic goals. By conditioning policies on this extended goal representation, our framework enables targeted, controllable scenario synthesis. We demonstrate that our approach successfully generates interactive driving behaviors that static-goal baselines cannot simulate, providing a flexible tool for long-tail scenario generation in AV testing.
\end{abstract}

\section{Introduction}
\label{sec:introduction}

Rigorous autonomous vehicle (AV) testing depends on being able to simulate long-tail driving scenarios with agents that react to the vehicle under test rather than ignore it.
Recent advances have shown that large-scale, multi-agent self-play can be used to train such agents~\cite{cusumano2025robust, cornelisse2025building, cornelisse2026humanlike, wu2026terrazero, qiu2026heterogeneous, wang2026learning}.
In contrast, the common alternatives for simulation agents (log replay, hand-scripted actors, and rule-based car-following models \cite{treiber2000congested}) fail to produce realistic interactive behavior.

Gigaflow \cite{cusumano2025robust} is the clearest demonstration of this, where copies of one policy control every traffic participant.
This policy is trained against copies of itself in large-scale simulation to optimize a parameterized reward function, with each agent receiving different parameters sampled during training.
Agents observe a local view of the map, goal locations, their own state, nearby agents, and their reward parameters.
From this setup, and without any human driving data, the policy learns generalist, robust, and human-like driving.
Conditioning on the reward parameters allows the agent's behavior to be controlled at inference time (e.g. to create harder simulations by making agents more aggressive).

Simulation agents are used to test an AV system.
During validation, challenging agent interactions such as cut-offs, tailgating, and brake checks are of particular interest; nominal driving behavior is simply not very useful for testing advanced AV systems.
What we want is an agent that can reliably produce these interactions in simulations with the AV driving system.
The standard self-play recipe, however, cannot achieve this.
As demonstrated below, the reward parameters and static goals used to condition these agents are insufficient to get the agent to perform these key interactions with another agent.
This begs the question: how can we modify the multi-agent self-play training recipe to add support for agent interactions while preserving all the positive attributes of this recipe?

\begin{figure}[t]
    \centering
    \includegraphics[width=\linewidth]{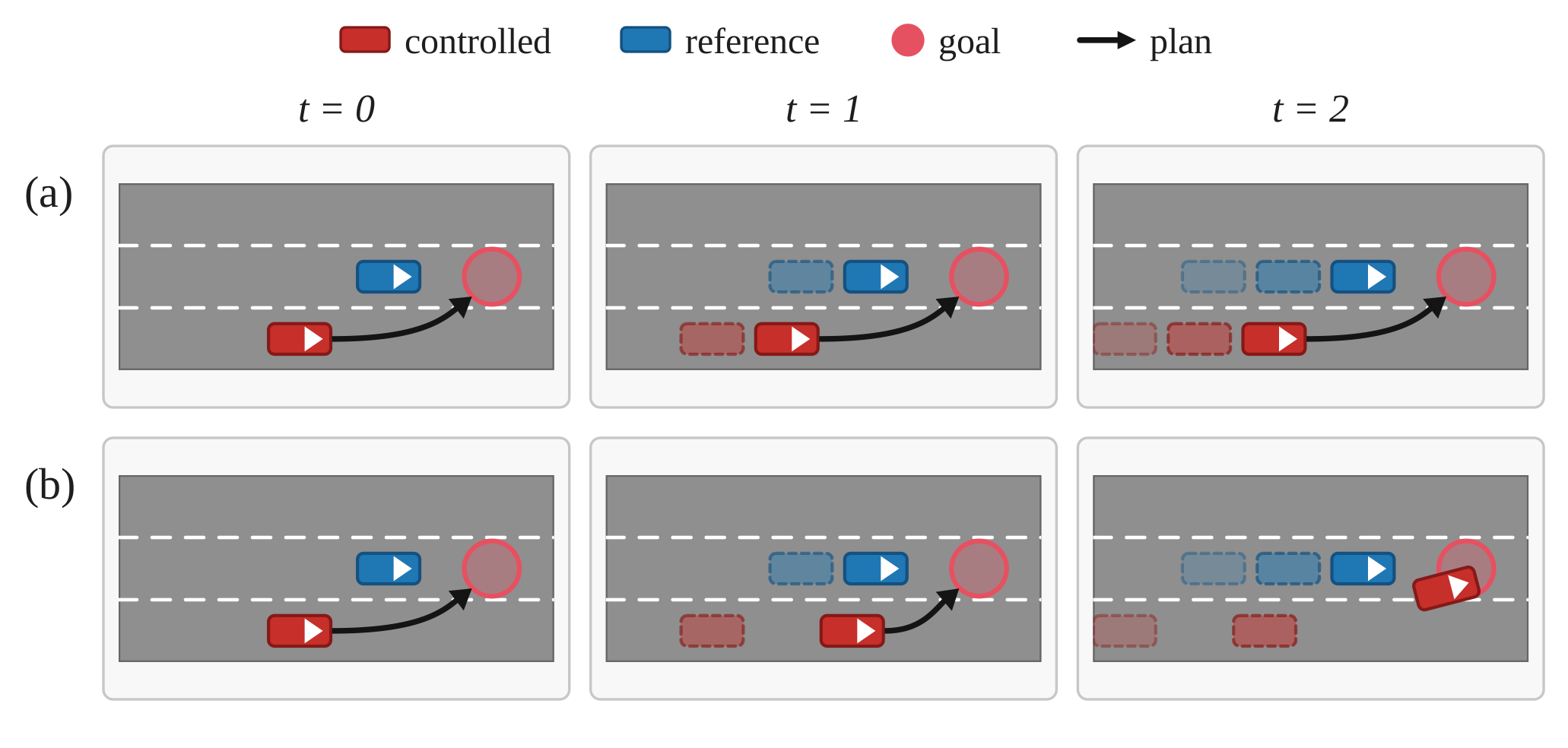}
    \caption{A goal in front of the blue vehicle moves with it (dashed boxes:
    earlier poses). The arrow represents the red agent's plan: a lane change
    to reach the \emph{current} goal location.
    (a)~Existing self-play agents train with only stationary goals. When presented with a moving goal at inference time, the agent keeps making the same plan and never actually reaches the goal.
    (b)~By adding moving goals to the self-play training recipe, the red agent learns to accelerate and overtake the blue agent to reach the moving goal. This lets us reliably create cut-off scenarios. Different goal placements can be used to create other interactions (e.g. tailgating, an agent in a blind spot).}
    \label{fig:1}
\end{figure}

In this work, we improve the multi-agent self-play recipe of Gigaflow by adding support for agent-relative goals through an extended goal interface.
These agent-relative goals are defined with respect to another reference agent and therefore move along with that reference agent.
These new goals allow us to simulate specific agent interactions that otherwise cannot be expressed with the previous static goals.
For example, we can command a red agent to cut-off the blue agent by placing the red agent's goal in front of the blue agent (\Cref{fig:1}).
Because the goal is agent-relative, it moves with the blue agent.
The only way for the red agent to reach its goal is to get in front of the blue agent.

We implement this extension with the smallest change possible: an updated goal sampling mechanism during training and a single bit added to the agent observation.
We find that this is sufficient to achieve the desired interactive behaviors.

Specifically, our contributions are:
\begin{enumerate}[label=(\roman*)]
    \item we introduce agent-relative goals as a novel extension to the multi-agent self-play paradigm
    \item we show a set of scenario demonstrations enabled by agent-relative goals
    \item we demonstrate that adding agent-relative goals preserves the original baseline behavior 
\end{enumerate}
%

%

\section{Related Work}
Autonomous driving simulation traditionally relies on rule-based car-following and lane-changing models~\cite{kesting2007general,kesting2010enhanced}, open-loop log-replay frameworks and data-driven simulators~\cite{gulino2023waymax,li2022metadrive,vinitsky2022nocturne}, or learned deep generative models for multi-agent trajectory prediction and scenario synthesis~\cite{huang2024versatile,ivanovic2019trajectron,suo2021trafficsim,xu2022bits,zhong2023guided}. To author explicit and programmatic test conditions, domain-specific probabilistic scenario description languages like Scenic~\cite{fremont2019scenic,fremont2023scenic} allow one to formally define spatial constraints, dynamic behaviors, and scene configurations. Alternatively, building on the success of multi-agent self-play reinforcement learning (RL) across complex domains~\cite{openai2021asymmetric,silver2017mastering,vinyals2019grandmaster}, recent driving simulators apply self-play RL at scale to learn interactive, closed-loop traffic behaviors~\cite{feng2023dense,zhang2023learning}. Most notably, Gigaflow~\cite{cusumano2025robust} demonstrates that massive-scale multi-agent self-play conditioned on static map goals yields robust, human-like driving policies without requiring human driving demonstrations. However, existing self-play frameworks like Gigaflow lack interfaces for fine-grained dynamic control such as agent-relative positional goals or explicit speed constraints, making it difficult to reliably synthesize specific safety-critical edge cases on demand.
In this work, we address this limitation by extending the multi-agent self-play framework with speed-constrained and agent-relative goal representations, enabling controllable and targeted synthesis of interactive, safety-critical driving scenarios for AV testing.

\section{Extended Multi-Agent Self-Play Framework}

In this section we describe the baseline multi-agent self-play system based on Gigaflow~\cite{cusumano2025robust} and our novel extension to add agent-relative goals.
This new type of goal enables new behaviors from self-play agents.

\subsection{Baseline System}
\label{sec:baseline}
We build on the goal-conditioned multi-agent self-play framework of
Gigaflow~\cite{cusumano2025robust}.
A single parametric policy controls every traffic participant and is trained against copies of itself in large-scale multi-agent simulation, with no scripted scenarios and no human driving data.
The reward function is parameterized (e.g. the coefficients for various reward terms), with the reward parameters for each agent randomized during training and given as input to the agent during the episode.
Each agent observes a local view of the map, its own state, nearby agents, and its own goals and conditioning; it does not observe other agents' goals or conditioning, so it has to infer intentions of other agents from their motion alone.
The reward is a sparse bonus for reaching goals plus additional terms to penalize unsafe and uncomfortable driving, similar to Gigaflow~\cite{cusumano2025robust}.
Training uses distributed proximal policy optimization (PPO)~\cite{schulman2017proximal, wijmans2020ddppo}.

A goal is given as a tuple $(p, \delta)$, where $p$ denote a position on the road graph and $\delta$ is an acceptance radius.
An agent \emph{reaches} a goal when its center position lies within $\delta$ of $p$.
Each agent is given a chain of such goals. When the agent reaches the first goal in the chain that goal is removed and a new goal is appended to the end of the chain.
During training, these goal positions are sampled based on the road topology.
Crucially, each goal position $p$ is a static point on the map.
Below, we show that this goal representation is inadequate to encode agent interactions and present a novel extension to address this limitation.

Gigaflow also supports having a final goal that requires the agent's speed to be below 3 m/s (in addition to having a center position within $\delta$ of the goal position).
While such speed constraints can be useful to express certain behaviors (e.g. parking), they are insufficient to express behaviors such as cut-offs or tailgating.
Since these interactions do not have a natural ``final'' state (e.g. tailgating is an ongoing behavior), we focus on waypoint-style goals that do not impose a speed constraint in this work.
Speed constraints can be easily added to agent-relative goals to enable additional behaviors (e.g. to make an agent get in front of the reference agent and then break).

\begin{figure}[t]
    \centering
    \includegraphics[width=\linewidth]{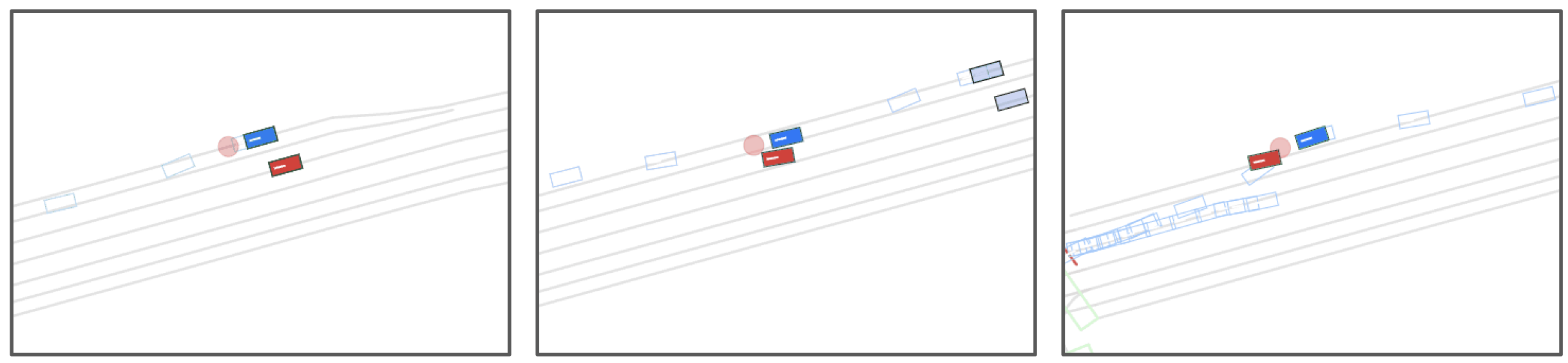}
    \caption{\textbf{Cut-off:} The red agent receives a goal 5 meters in front of the blue agent (left); the red agent overtakes the blue agent (middle); the red agent reaches its goal (right).}
    \label{fig:cutoff}
\end{figure}

\begin{figure}[t]
    \centering
    \includegraphics[width=\linewidth]{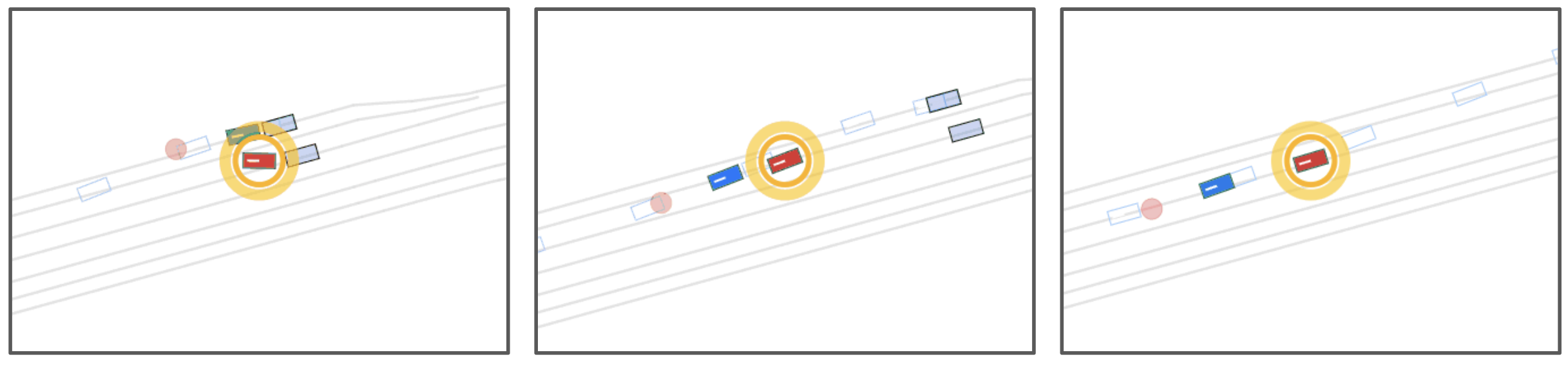}
    \caption{\textbf{Failure of baseline model:} Given the same simulation, an agent trained with only static goals simply follows behind the blue agent, never reaching the goal.}
    \label{fig:static-goal-fail}
\end{figure}

\subsection{Agent-Relative Goals}
\label{sec:relative}

Adversarial AV scenarios are often defined relative to the vehicle under test: cut in front of it, tailgate it, hold a position beside it, brake check it.
A static goal cannot express these, because where the position of the vehicle under test is a priori unknown and interactions such as tailgating require tracking a position that moves over time.
Placing a static goal at the spot where a cut-off should occur does not produce a cut-off: the agent can simply wait for the vehicle under test to pass by, and then go to the goal position.
We therefore need goals that are anchored to another agent and move with it, such as ``ten meters in front of that vehicle''.

To address this, we introduce agent-relative goals, i.e., goals defined with respect to a reference agent.
Each agent-relative goal is defined as a tuple $(a, q, \delta)$, where $a$ is the reference agent, $q$ is a position offset relative to the reference agent, and $\delta$ is an acceptance radius.
At a given time $t$, the position of the goal in global coordinates is $p_t = T^\textrm{world}_{a(t)} q$, where $T^\textrm{world}_{a(t)}$ is the coordinate transformation from the reference frame of agent $a$ at time $t$ (i.e. with the origin at the agent's center and aligned with the agent's heading) to the global reference frame.
The norm of the position offset $q$ is bounded during both training and inference (e.g. at most 2 lane widths laterally).

We need to define a unified observation space for both static and agent-relative goals.
To keep the static goal observation as similar to the baseline as possible, we want an observation space to be the current goal position, the acceptance radius, and a third signal that is only present for agent-relative goals.
There are many options for what this could be: we could encode the agent ID of reference agent, the velocity of the reference agent, or the agent-relative offset $q$.
We choose to use the minimal possible signal: a single bit saying whether the goal is static or agent-relative.
For a static goal $(p, \delta)$, the goal observation is $(p, \delta, 0)$.
For an agent-relative goal $(a, q, \delta)$, the goal observation at time $t$ is $(p_t, \delta, 1)$.
While this representation does not explicitly encode which agent is the reference agent, we find that in practice it is sufficient for the model to learn how to reach these goals.
We leave an exploration of alternative representations to future work.

\begin{figure}[t]
    \centering
    \includegraphics[width=\linewidth]{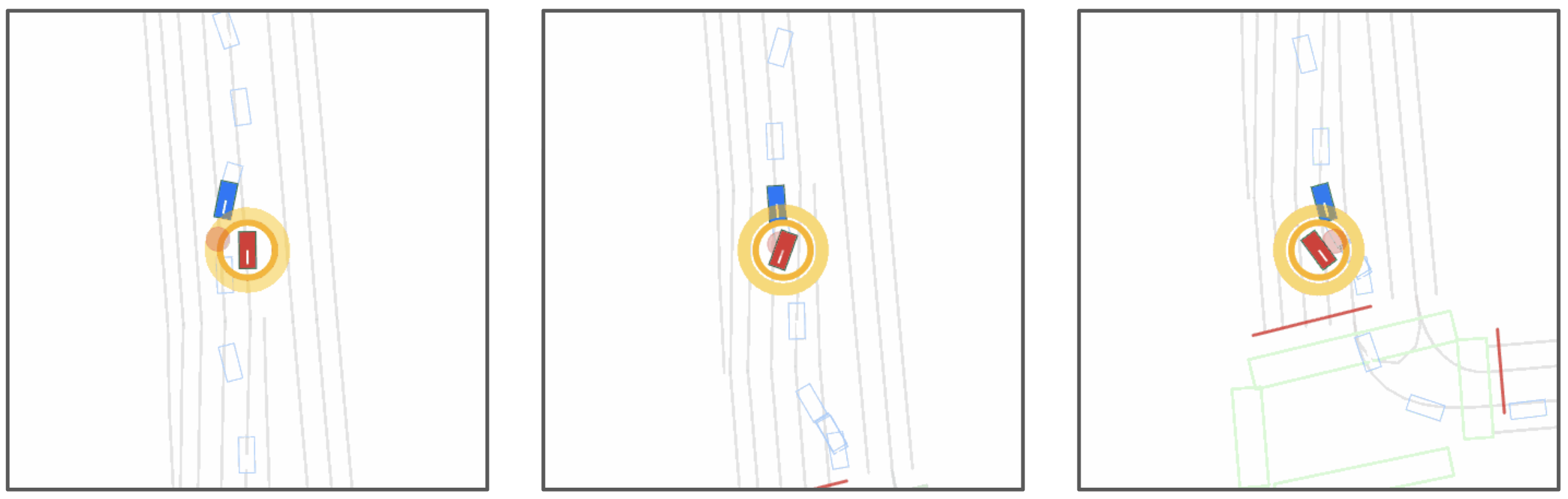}
    \caption{\textbf{Brake checking:} The red agent cuts 5 meters in front of the blue agent, forcing the blue agent to slow down and swerve to avoid a collision.}
    \label{fig:brake-check}
\end{figure}

\begin{figure}[t]
    \centering
    \includegraphics[width=\linewidth]{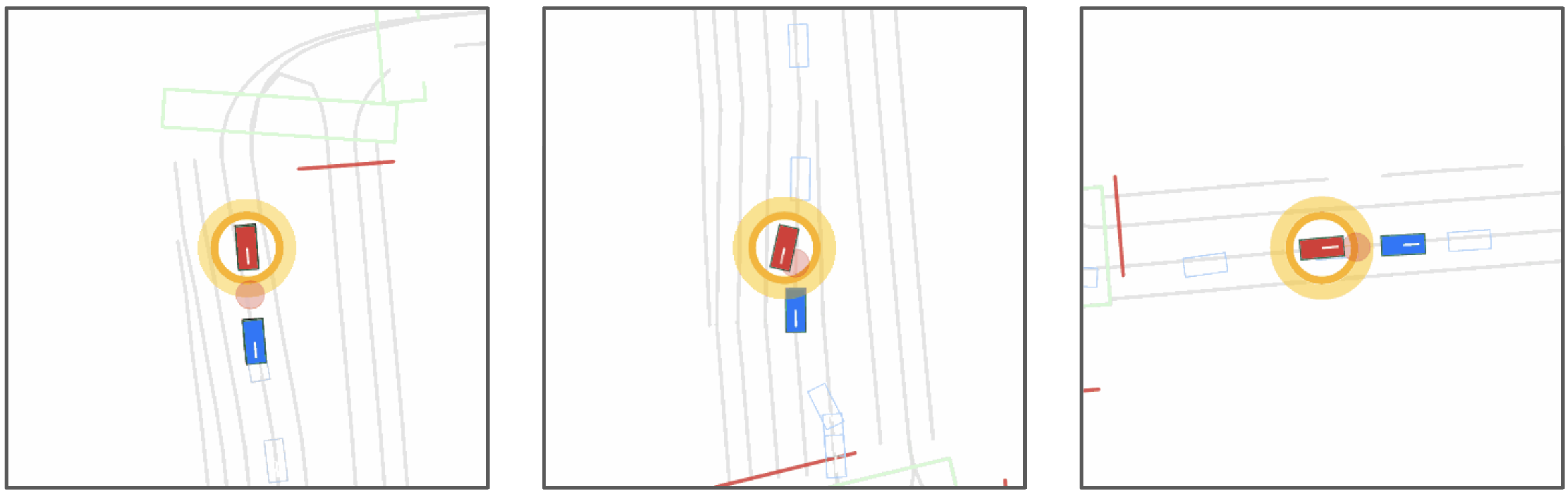}
    \caption{\textbf{Tailgating:} The red agent tailgates the blue agent 5 meters behind.}
    \label{fig:tailgating}
\end{figure}

\begin{figure}[t]
    \centering
    \includegraphics[width=\linewidth]{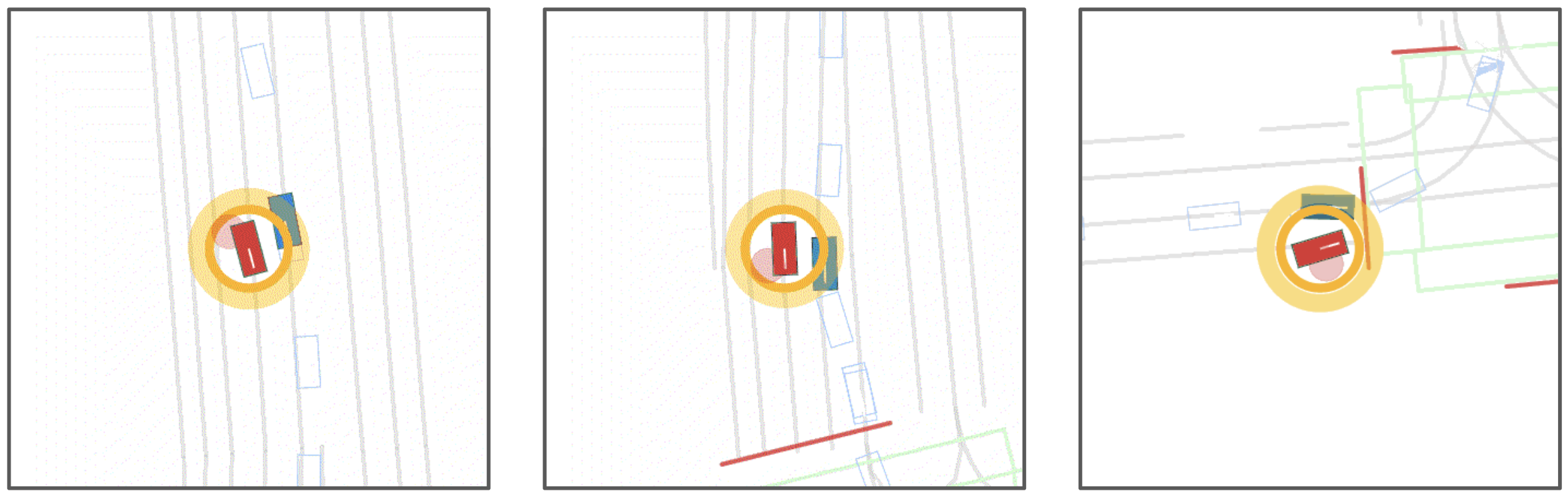}
    \caption{\textbf{Side-by-side driving:} The red agent remains 5 meters to the right of the blue agent. This can be used to test the blue agent's ability to lane change in dense traffic.}
    \label{fig:on-right}
\end{figure}

\subsection{Training}
\label{sec:training}

Given a model trained using the baseline system described in Sec.~\ref{sec:baseline} (i.e. with only static goals), we can perform inference using agent-relative goals by simply dropping the binary flag from the goal representation.
However, this is insufficient: the behaviors required to reach static goals are different from those required to reach agent-relative goals, and the agent must learn this by seeing agent-relative goals during training.
For example, consider an agent-relative goal placed in front of the reference vehicle.
Given a static goal in front of the reference vehicle, the easiest way to reach the goal is to wait for the reference vehicle to pass and then drive to the goal.
In contrast, an agent-relative goal in front of the reference vehicle will keep moving, and reaching the goal requires overtaking the reference agent.
If the agent treats the agent-relative goal as static, it will consistently tail behind the reference vehicle, all the while thinking it is close to reaching the goal.
To reliably reach agent-relative goals, an agent must be trained with such goals to learn the novel behaviors these goals require.

During training, we sample a mixture of static and agent-relative goals with some mixture probability, drawn per agent at the start of an episode, so that a fraction of the agents pursue agent-relative goals throughout.
A fixed probability $\lambda$ determines whether a given agent will receive static or agent-relative goals.
Setting $\lambda = 0$ recovers the baseline training with only static goals.
To sample agent-relative goals, we sample the reference agent from the set of nearby agents and sample an offset $q$ using a configurable distribution.
Similar to static goals, when an agent reaches its agent-relative goal a new goal is sampled.

During inference, the reference agent is pinned to a chosen participant instead of being sampled, typically the vehicle under test.
Offsets are part of the goal definition, so different agents in the scenario can be given different relations to various reference agents.

\section{Experiments}

We now present scenario families enabled by the two extensions.
Each scenario is produced by a single policy that controls every participant, and is specified by given goals and initial states.
There are no scripted trajectories and no per-scenario reward tuning, so the scenarios present reactive agent behaviors.

\subsection{Implementation}

We implement the system for training with agent-relative goals as described in Sec.~\ref{sec:training}.
We simplify the distribution over agent-relative offsets $q$ by constraining the offsets to the four cardinal directions relative to the reference agent: front, back, left, and right.
A categorical distribution assigns probabilities to each cardinal direction, and a 1D distribution describes the distance along that direction for the offset.
We use a distribution with a larger range of offsets for the longitudinal directions compared to the lateral directions.
The binary flag in the goal representation is concatenated with the existing goal embedding that is input to the model.
The policy is trained with 10\% agent relative goals and 90\% static goals ($\lambda = 0.1$), whereas the baseline policy is trained and tested solely on static goals ($\lambda = 0$).

\subsection{Impact on Existing Capabilities}

To evaluate the impact agent-relative goals have on overall driving ability, we focus on two key metrics.
The collision percentage measures the percentage of agents that collide with another agent at any point during a self-play rollout.
The off-road percentage measures the percentage of timesteps in which an agent's center position is outside the drivable region of the map during self-play rollouts.
These metrics are evaluated in self-play using only static goals ($\lambda = 0$).
To solely analyze the effect of the proposed goal representation on the key metrics, we present the differences between these percentages for the policy with agent-relative goals and the baseline policy.



The new policy with agent-relative goals decreases the collision percentage by 0.03\% and the off-road tick percentage by 0.01\% compared to the baseline.
These results show that agent-relative goals seamlessly extend existing self-play infrastructure without compromising baseline policy performance on key metrics.
Adding agent-relative goals does not degrade the existing agent behavior when using static goals.
Qualitatively, we observe no difference in behavior between the baseline policy and the new policy when evaluated with static goals.

\subsection{Case Studies}

\Cref{fig:cutoff} shows a cut-off scenario.
The red agent is given a single goal 10 meters in front of the blue agent, and starts behind it in an adjacent lane.
The specification says only \emph{in front of that agent, at that distance}, so the lane selection, the gap acceptance, and the timing of the merge are all left to the policy.
Because the goal is relative to the blue agent, the red agent is chasing a target that moves forward as the blue agent drives, which forces it to close the gap before it can adversarially cut-off.
Note that without a binary flag indicating that the given goal is an agent-relative one, the red-agent does not catch-up to the moving goal target, because it cannot distinguish the goal from a static goal in that case, highlighting the need for this additional bit of information.
Similarly, the baseline model trained with only static goals also fails to reach the goal, as shown in \Cref{fig:static-goal-fail}.

In \Cref{fig:brake-check}, a goal held in front of the blue agent produces brake checking, where the red agent takes the position immediately ahead and holds it, forcing the blue agent to manage a close lead vehicle.
In \Cref{fig:tailgating}, a goal held behind the blue agent produces tailgating, where the red follows persistently and closes back in whenever the blue agent pulls away (e.g. after a stop sign).
In \Cref{fig:on-right}, a goal held to the right of the blue agent produces side-by-side driving, with the red agent tracking a position off the blue agent's side.
In all cases the red agent attempts to maintain the imposed relation as the blue agent changes speed.

\begin{figure}[t]
    \centering
    \includegraphics[width=\linewidth]{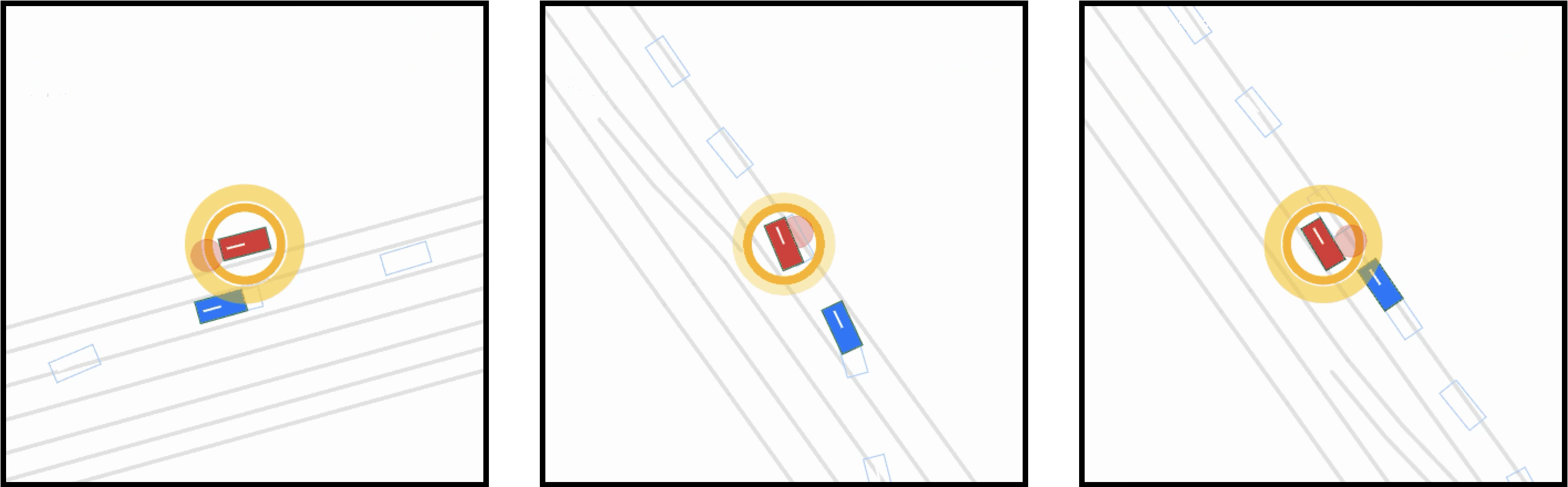}
    \caption{\textbf{Cut-off and brake check:} The red agent moves toward a relative goal 5 meters to the right of the blue agent (left); it moves toward its next goal 10 meters ahead of the blue agent to execute a cut-off maneuver (middle); the red agent decelerates to reach a goal 5 meters in front of the blue agent, performing a brake check (right).}
    \label{fig:multi_cutoff}
\end{figure}

\Cref{fig:multi_cutoff} composes these relations over time.
The red agent is given a chain of three agent-relative goals against the same reference agent: 5 meters to its right, then 10 meters ahead of it, then 5 meters ahead of it.
The first two stages specify a more fine-grained cut-off behavior than a single front-relative goal, because the red agent establishes a lateral position before it commits longitudinally.
At the third stage, since the goal tracks the blue agent's current pose, shrinking the requested gap from 10 meters to 5 meters while staying in front of it can only be satisfied by decelerating, so a brake check behavior emerges.

\subsection{Quantitative Evaluation}

We measure the agent's ability to reliably produce interaction behaviors on a set of test cases.
Each scenario is rolled out 1000 times from a fixed initial state sequence of agent-relative goals on  two maps (one from San Francisco and one from Austin).
Actions are sampled stochastically from the policy.
Both the baseline policy without agent-relative goals and the updated policy with agent-relative goals receive the identical goals, except that the baseline policy doesn't condition on the binary flag since it only saw stationary goals during training.
Success is measured by whether the agent reaches all the agent-relative goals in a fixed time window.

\textsc{cutoff} uses a single agent-relative goal in front of the reference agent, with the ego agent starting behind the reference agent, as in Fig.~\ref{fig:cutoff}.
\textsc{cutoff\_and\_slow\_down} uses a sequence of three agent-relative goals: on the side of the reference agent, in front of the reference agent with a large offset, and in front of the reference agent with a small offset.
\textsc{leading} uses a single agent-relative goal ahead of the reference vehicle, with the ego agent starting ahead of the reference agent as well.
\textsc{on\_right} uses a single agent-relative goal to the right of the reference agent, as in \Cref{fig:on-right}.
\textsc{tailing} uses a single agent-relative goal behind the reference agent, as in \Cref{fig:tailgating}.

\Cref{tab:agent-relative-goal-success-maps} reports the results.
The two cut-off scenarios separate the policies
sharply.
The agent-relative policy overtakes and merges ahead of the reference
vehicle in 79.0--84.2\% of rollouts and completes the three-stage chain in
70.7--84.7\%, while the baseline, which treats the goal as a static point,
succeeds in at most 14.6\% and essentially never completes the chain; its
occasional successes occur when the reference vehicle is slow or stopped and the
goal briefly stops moving.
The three additional rows are less discriminative by construction: the agent starts about 5\,m from its relative goal, and a single pass through the point satisfies the criterion.
In \textsc{leading} the point 5\,m ahead of the reference vehicle sweeps over any
agent that merely stays in its lane as the reference closes the gap, which is why
the baseline also scores 100\% there. Even under this lenient criterion the
baseline is less reliable when the relation requires a lateral move
(\textsc{on\_right}, 72.8--81.6\%) or catching a receding target from behind
(\textsc{tailing}, 91.5--94.8\%), whereas the agent-relative policy establishes
every relation in at least 98.2\% of rollouts.
These results highlight the fact that training with agent-relative goals is crucial for non-trivial adversarial cut-off behaviors.

\begin{table}[t]
\centering
\caption{Success rate for interaction behaviors.}
\label{tab:agent-relative-goal-success-maps}
\begin{tabular}{lrrrr}
  \toprule
  & \multicolumn{2}{c}{San Francisco} & \multicolumn{2}{c}{Austin} \\
  \cmidrule(lr){2-3} \cmidrule(lr){4-5}
  With agent relative goals? & Yes & No & Yes & No \\
  \midrule 
  \textsc{cutoff}                  & 84.2\%  & 14.6\%  & 79.0\%  & 1.0\% \\
  \textsc{cutoff\_and\_slow\_down} & 70.7\%  & 3.7\%   & 84.7\%  & 0.1\% \\
  \textsc{leading}                 & 98.2\%  & 100.0\% & 99.8\%  & 100.0\% \\
  \textsc{on\_right}               & 100.0\% & 81.6\%  & 100.0\% & 72.8\% \\
  \textsc{tailing}                 & 100.0\% & 94.8\%  & 99.7\%  & 91.5\% \\
  \bottomrule
\end{tabular}
\end{table}

\section{Conclusion}

We extend goal-conditioned multi-agent self-play with agent-relative goals.
This addition is opt-in, introducing minimal interface changes without altering the core reward structure or training algorithm.
Training on a mixture of static and agent-relative goals yields a single controllable policy capable of executing novel behaviors while maintaining baseline capabilities.
Consequently, these lightweight additions yield more controllable policies capable of executing a broader spectrum of scenarios relevant to AV testing.


\bibliographystyle{IEEEtran}
\bibliography{root} 

\end{document}